\documentclass[sigconf,authorversion,nonacm]{acmart}
\usepackage{subfigure}
\usepackage{acronym}
\usepackage{comment}
\usepackage{enumitem}
\usepackage{graphicx,color,soul}
\usepackage{acronym}
\usepackage{makecell}
\setstcolor{red}

\usepackage{amsmath}
\usepackage{graphicx}
\usepackage{caption}

\begin{document}

\title{Tailoring Graph Neural Network-based Flow-guided Localization to Individual Bloodstreams and Activities}

\author{Pablo Galván, Sergi Abadal}
\affiliation{
  \institution{Universitat Polit\`ecnica de Catalunya}
  \city{Barcelona}
  \country{Spain}}
\email{{name.surname}@upc.edu}

\author{Filip Lemic\footnotemark\authornote{Corresponding author.}, Gerard Calvo, Xavier Costa Pérez\footnotemark\authornote{X. Costa is also with NEC Labs Europe GmbH, Germany and ICREA, Spain.}}
\affiliation{
  \institution{i2Cat Foundation}
  \city{Barcelona}
  \country{Spain}}
  \email{{name.surname}@i2cat.net}

\renewcommand{\authors}{P. Galván, F. Lemic, G. Calvo Bartra, S. Abadal, X. Costa Pérez}
\renewcommand{\shortauthors}{P. Galván, F. Lemic, G. Calvo, S. Abadal, X. Costa}

\begin{abstract}

Flow-guided localization using in-body nanodevices in the bloodstream is expected to be beneficial for early disease detection, continuous monitoring of biological conditions, and targeted treatment. 
The nanodevices face size and power constraints that produce erroneous raw data for localization purposes. 
On-body anchors receive this data, and use it to derive the locations of diagnostic events of interest. 
Different \ac{ML} approaches have been recently proposed for this task, yet they are currently restricted to a reference bloodstream of a resting patient. 
As such, they are unable to deal with the physical diversity of patients' bloodstreams and cannot provide continuous monitoring due to changes in individual patient's activities. 
Toward addressing these issues for the current \ac{SotA} flow-guided localization approach based on \acp{GNN}, we propose a pipeline for \ac{GNN} adaptation based on individual physiological indicators including height, weight, and heart rate.
Our results indicate that the proposed adaptions are beneficial in reconciling the individual differences between bloodstreams and activities. 
\end{abstract}

\maketitle

% \fi
%\IEEEpeerreviewmaketitle

%!TEX root = main.tex

\acrodef{ML}{Machine Learning}
\acrodef{THz}{Terahertz}
\acrodef{GNN}{Graph Neural Network}
\acrodef{ZnO}{Zinc Oxide}
\acrodef{IMU}{Inertial Measurement Unit}
\acrodef{RF}{Radio Frequency}
\acrodef{SINR}{Signal to Interference and Noise Ratio}
\acrodef{NN}{Neural Network}
\acrodef{SotA}{State-of-the-Art}
\acrodef{HGT}{Heterogeneous Graph Transformer}
\acrodef{GMM}{Gaussian Mixture Model}
\acrodef{CHLR}{Central High-speed Lane Routing}
\acrodef{SDM}{Software-Defined Metamaterial}
\acrodef{GAT}{Graph Attention Network}
\acrodef{GCN}{Graph Convolutional Network}
\acrodef{BVS}{BloodVoyagerS}
%!TEX root = main.tex

\section{Introduction}

The introduction of in-body nanodevices to the field of medicine has a potential to revolutionize medical applications by enabling early diagnostics, continuous monitoring, and precise interventions within the human bloodstreams~\cite{ohshiro2022nanodevices,doi:10.1021/nn406285x,mukherjee2020nanodevices}. 
The nanodevices are envisioned to be passively flowing in the bloodstream, periodically sensing for biological indicators of interest, and communicating their findings to on-body anchors for further processing.
Assigning location indicators to the sensed events, which is the main proposition of flow-guided localization, is expected to further enhance the diagnostic, monitoring, and treatment possibilities~\cite{lemic2021locationaware}.

\begin{figure}
    \centering
    \captionsetup{type=figure}
    \includegraphics[width=\linewidth]{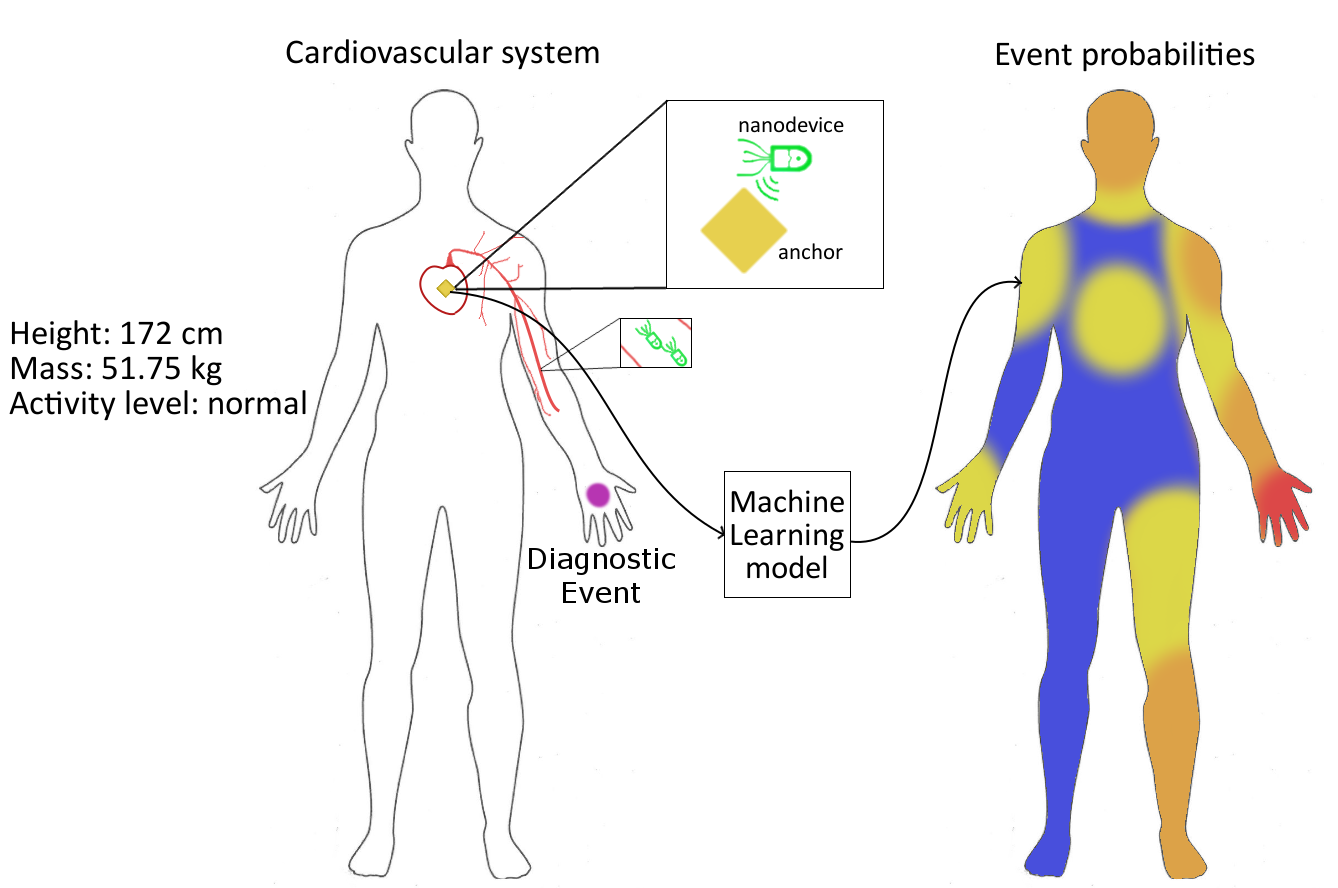}
    % \vspace{-2mm}
    \captionof{figure}{High-level overview of flow-guided nanoscale localization tailored to individual profiles}
    \label{fig:intro}
    \vspace{-3mm}
\end{figure}

Flow-guided nanoscale localization is a technique for locating an event of medical interest using nanodevices within the bloodstream, without requiring the nanodevices to determine their own location~\cite{pascual2023analytical}, as indicated in Figure~\ref{fig:intro}. 
In contrast, as a minimum requirement the nanodevices are envisioned to communicate their event detection indicators to an on-body anchor positioned in the proximity of the heart, as the heart represents the only bodily region in which the nanodevices pass in each of their iterations through the bloodstream.
The communication is usually assumed to be done using high-frequency pulse-based wireless communication in the \acf{THz} frequency band to meet the stringent constraints on the physical size of the nanodevices~\cite{abadal2015time,hossain2019stochastic}. 

In-body nanodevices face operational constraints due to their physical size and energy limitations, as well as due to their erratic movement in the bloodstream. 
Thus, the communication with the anchor is generally erroneous~\cite{bartra2023graph}, while an event might not be detected by a nanodevice, despite the fact that the nanodevice passed through the cardiovascular path containing the event. 
For this reason, flow-guided localization is a challenging task to achieve with traditional algorithms, resulting in several proposals that utilize \acf{ML} for this purpose~\cite{bartra2023graph,torres,simonjan2021inbody}. 

While these solutions have yielded promising performance in a reference bloodstream of a resting patient, none of them considered tailoring the approaches to individual bloodstreams and activities. 
Such considerations are necessary for seamless administration of flow-guided localization systems across patients, as well as for enabling continuous (i.e., activity agnostic) localization within the bloodstreams.
Understanding how individual factors such as weight, height, and heart rate affect the localization accuracy is crucial for optimizing the performance of such systems. 

Toward addressing these concerns, we explore how variations in patients' physiological indicators affect the performance of a contemporary flow-guided nanoscale localization approach based on \acfp{GNN}~\cite{bartra2023graph}. 
The reference \ac{GNN} model is extended so that it utilizes biological indicators including weight, height, and heart rate as attributes for the \ac{GNN} adaptation.
Nine individual profiles have been defined and employed to evaluate the impact of profile adaptations on localization accuracy.
Our results indicate that the proposed extensions to the \acf{SotA} \ac{GNN}-based flow-guided localization approach can enhance the adaptability of the model, suggesting that its operability can be extended beyond the reference bloodstream of a resting patient toward individual bloodstreams and activities. 

This work is built on top of the study conducted by Calvo~\emph{et al.}~\cite{bartra2023graph}, which leverages \acp{GNN} for flow-guided localization while accounting for communication and energy limitations at the nanodevice level. Specifically, our research includes the following contributions:
\begin{itemize}
    \item Addition of a master node to the graph structure, holding information about the body shape and heartbeat rate.
    \item Improvement of edge weights in the input graph design by including the probability of a region being visited by the nanodevices in a single iteration through the bloodstream.
    % \item A pipeline capable of generating simulation data for different body types and heartbeat rates.
    \item Performance improvement on most body types and activity levels in comparison to the original \ac{GNN} design from~\cite{bartra2023graph}.
\end{itemize}

%!TEX root = main.tex

\section{Related Works}

The promise of in-body localization has been demonstrated in~\cite{10.1145/3230543.3230565}, where the authors propose a backscatter-based in-body communication and localization system. The system is experimentally shown to yield centimeter-level accuracy in human phantoms and animal tissue. The initial promise encouraged further research on downscaling the physical size of in-body devices to the nanoscale, paving the way toward the administration of in-body nanodevices into patients' bloodstreams. Their physical sizes have to be comparable to or smaller than those of the red blood cells, or around 5 microns. Localization of such devices in the bloodstream is coined under the term flow-guided nanoscale localization. 

Several approaches have been proposed for flow-guided localization of events of diagnostic interest.
The proposal in~\cite{lemic2021locationaware} envisions using \acp{SDM} as on-body anchors and multi-hopping-based wireless communication for localizing in-body nanodevices.
While this approach promises high localization accuracy, it requires a large number of nanodevices and their complex functionalities such as the support for multi-hopping communication. 
This is currently infeasible in practice, primarily due to the operational constraints at the nanodevice level. 

Moreover, the approach proposed in~\cite{simonjan2021inbody} leverages conceptual nanoscale \acp{IMU} to measure nanodevices' acceleration and rotation. This data is sent to an anchor near the heart and used to determine the location of the nanodevices over time. After communicating the data to the anchor, the \acp{IMU} are reset to prevent error accumulation. While this approach shows encouraging results, using \acp{IMU} adds complexity and introduces challenges to nanodevice fabrication, integration, and data storage.

The approach in~\cite{torres} envisions the nanodevices to provide their travel concentration level to the anchor near the heart. 
This data is then clustered using a k-means algorithm, where each group corresponds to a different closed loop in the circulatory system. These clusters are fed to a decision tree that classifies the data into different bodily regions, providing region-level location estimation. 
This is followed by utilizing an \ac{ML} model to determine the transition probabilities in a Markov chain. 

In addition, the approach in~\cite{bartra2023graph} leverages \acp{GNN} to localize diagnostic events in the cardiovascular system. The circulation times from different nanodevices are reported to an anchor, which aggregates them to create features for the \acp{GNN} model from which the event location is predicted.
In particular, \acp{HGT} are used, allowing for multiple types of nodes, i.e., anchor and region nodes. 
Each anchor is connected to all regions, and each region is connected to its neighbor regions based on their structure in the human bloodstream. 
The authors have demonstrated that the \ac{GNN} approach yields higher localization accuracy and offers an extended coverage compared to the alternative approaches and is, therefore, selected as a baseline for this study. 

Finally,~\cite{pascual2023analytical} provides an analytical model of raw data for flow-guided localization. 
The model's outputted data is shown to be highly comparable to the corresponding data generated by a flow-guided localization approach. 
The approach from~\cite{pascual2023analytical} is envisioned to be used for training flow-guided localization approaches in a non-invasive way, which in turn paves the way toward eventually deploying flow-guided localization in patients' bloodstreams. 
As such, it is envisioned to be utilized to train approaches from~\cite{torres,bartra2023graph}.

However, the related works have yet to consider the physiological differences between individual bloodstreams or the adaptability of approaches to different patient activities. Therefore, these approaches might be only suitable for a reference bloodstream and a resting patient, limiting flow-guided localization's seamlessness and deployability potential.

%!TEX root = main.tex

\section{Considered GNN Model}

\subsection{GNN-enabled Flow-guided Localization}
\label{sec:gnn_architecture}

\begin{figure}[!t]
\centering
\includegraphics[width=\linewidth]{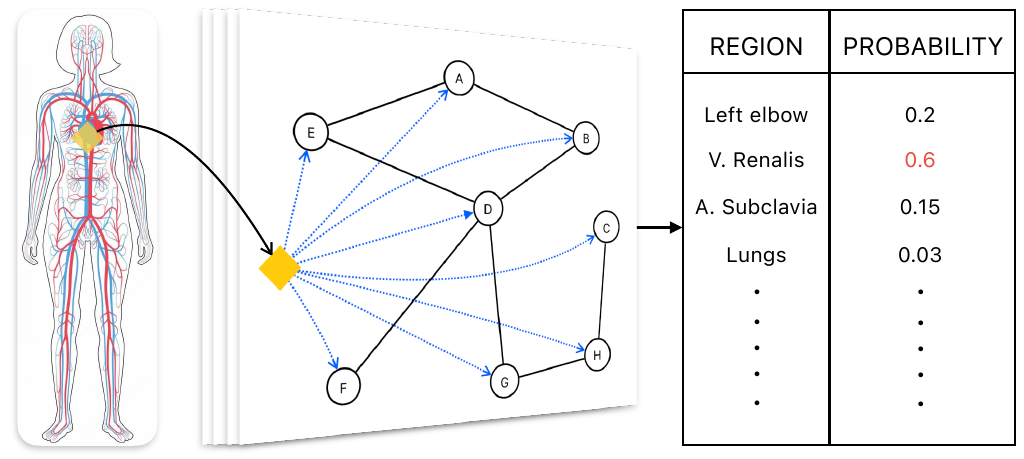}
% \vspace{-4mm}
\caption{Overview of the utilized GNN-based flow-guided nanoscale localization approach~\cite{bartra2023graph}}
\label{fig:vision}
\vspace{-2mm}
\end{figure}

As a reference model, we use the \ac{SotA} \ac{GNN} model from~\cite{bartra2023graph}, which acts as a pattern-matching function to locate the sensed events within the cardiovascular system given the set of observed signal features measured by the anchor.
Since the cardiovascular system is a structured and highly connected environment, it can be naturally represented as a graph (as shown in Figure~\ref{fig:vision}), which is why this approach holds significant potential for individualized flow-guided localization. 
In the graphs used in the reference study, each node corresponds to a body region of an arbitrary granularity, and the edges define the constraints for the movement of the nanodevices by linking adjacent regions in the cardiovascular system. 

The \ac{GNN} model utilizes region and anchor nodes. Region nodes hold the information of said regions, such as their type (e.g., organs, limbs or head, veins, arteries), length, and blood speed. 
Anchor nodes carry the information on the circulation times of the positive bits received from the nanodevices for the localization process. Because the raw data can be of varying length (the anchor receives an undefined number of positive event bits), the used \ac{GNN} expects the circulation time for positive event bits to be modeled as a \ac{GMM} with two Gaussian clusters. With this approach, the distribution parameters derived from the \ac{GMM} serve as features for the anchors alongside the average number of positive bits received, thus providing a fixed-length feature set for each anchor.

\subsection{GNN Architecture}

\begin{figure}[!t]
\centering
\includegraphics[width=\linewidth]{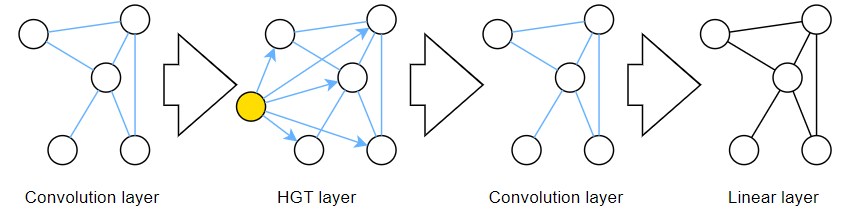}
% \vspace{-5mm}
\caption{Overview of the reference GNN architecture~\cite{bartra2023graph}}
\label{fig:gnn_architecture}
\vspace{-2mm}
\end{figure}

The original \ac{GNN} architecture is depicted in Figure~\ref{fig:gnn_architecture} and leverages the inherent structure of the graph representing the bloodstream. It is based on a \ac{HGT}, providing versatility to handle the system's multiple types of nodes, i.e., region nodes and anchors.
The \ac{GNN} starts by generating unique embeddings for each node type (applying a ReLU activation function to the initial node features, transforming the information into a latent space of higher dimensionality). 
Next, these embeddings are fed to an initial set of graph layers, which might be \acp{GAT}~\cite{velickovic2018graph} or \acp{GCN}~\cite{zhang2019graph} depending on the hyperparameter configuration, and it is followed by a suite of \ac{HGT} layers and a concluding convolution layer.

The initial convolution layers introduce non-linearity and adaptability into the model, aggregating information from each region node's neighbors depending on the weight between the nodes. In the case of \acp{GAT}, this weight is calculated as an attention score, a measure learned during training that assigns different relative importance to each node during the aggregation of features.
Following the initial layers, the \ac{HGT} layers deal with the complex interactions between different types of nodes.
These layers incorporate the information from the anchors to region nodes, enabling the model to capture the dynamic propagation of nanodevices through different body regions. By dynamically adjusting the importance of different nodes based on the information they carry, the HGT layers are able to provide a nuanced representation that encapsulates the spatial and temporal aspects of the nanodevices' propagation.

The model links the anchor node to all region nodes, eliminating the need for multiple stacked message-passing layers to ensure that information from the anchor reaches all region nodes. 
After the \ac{HGT} layers have processed the information, a final set of convolution layers is applied to refine the region nodes' representations. 
The architecture concludes with a final linear layer applied to the refined representations of the region nodes. 
The output then undergoes a sigmoid activation function to produce the final predictions, indicating the likelihood of an event occurring in each region.
This \ac{HGT} model provides initial probabilities for the occurrence of the detected event across all regions in the cardiovascular system. 

%!TEX root = main.tex

\section{GNN-based Localization Tailored to Individual Profiles}

\subsection{Extended Graph Designs}

The reference graph design was extended in several ways to improve the accuracy of the GNN on individual profiles. This included providing the GNN with information associated with each profile and making modifications to improve the structure itself.
Firstly, every extended design includes a master node connected to all other nodes (both region and anchor nodes). The basic version of this node uses the three attributes of each profile as features, i.e., the weight, the height, and the heart rate affecting the speeds of blood.

\begin{figure}[!t]
    \centering
    \includegraphics[width=0.8\linewidth]{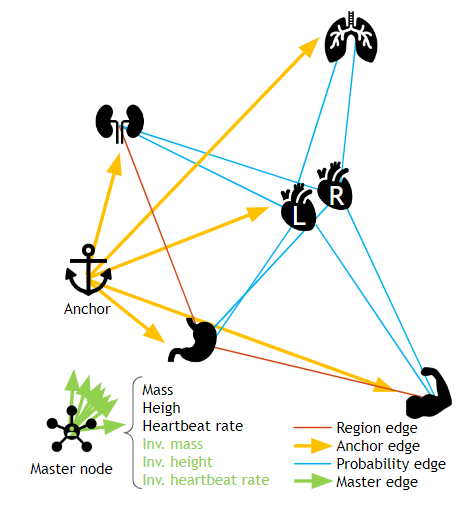}
    % \vspace{-3mm}
    \caption{Overview of the extended graph designs, showing the master node, its potential features, and the probability edges between the heart nodes and the other regions}
    \label{fig:extended_graphs}
    \vspace{-2mm}
\end{figure}

In addition to these features, the master node can hold the inverted value of these three attributes. These inverted values might benefit the model since increasing the weight and height might cause the nanodevices' iterations through the cardiovascular systems to require more time (since the vessels might be longer). 
Therefore, normalizing the circulation time by providing the \ac{GNN} with the inverse value of the features might help.

\begin{figure*}[!h]
    \centering
    \includegraphics[width=\linewidth]{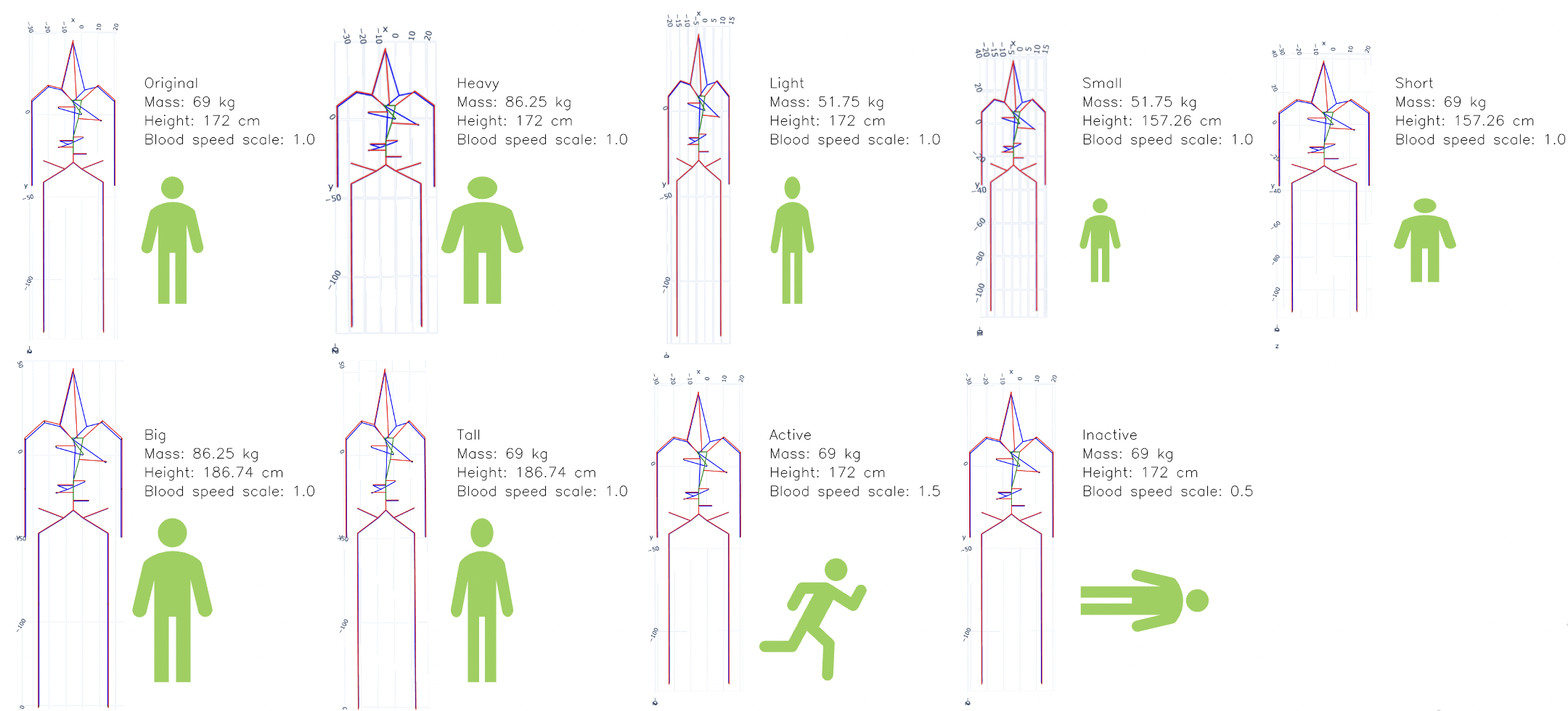}
    % \vspace{-2mm}
    \caption{Cardiovascular system for different patient profiles and their settings, with the size differences emphasized for visualization purposes}
    \label{fig:vasculature}
     \vspace{-2mm}
\end{figure*}

Besides the master node, a graph design with different edge types was tested. This new design connects the two heart region nodes (left and right heart) to all the other region nodes except for themselves. The weight of each of these new edges is the probability that the nanodevices visit the region adjacent to the heart node in one iteration. These probabilities were estimated through a Monte-Carlo search, using the evaluation setup outlined in Section~\ref{sec:results}. 
This design was considered as it allows the model to know which regions are more likely to be visited in an iteration, which allows it to infer more information regarding the path followed by the nanodevices.
As indicated in Figure \ref{fig:extended_graphs}, three combinations of these designs were tested and compared: a) graph with inverted master features only, b) graph with probability edges only and c) graph with probability edges and inverted master features.

These extensions allow for testing the model's performance and predict distinct realistic scenarios by knowing the physiological information of the patients.
We also assume exemplary activity levels changing the subjects' blood speeds.
Nonetheless, continuous activity recognition can be performed through their correlation subjects' heart rates (e.g.,~\cite{brage2007hierarchy}) or other physiological measurements, implemented at the anchor level. 
Using continuous values for these attributes instead of discrete classes associated with the tested profiles is expected to allow the model to estimate the locations of diagnostic events continuously across activities and for profiles beyond the scope of this study. 

\vspace{-1mm}
\subsection{Raw Data Transformation}

For evaluating the \ac{GNN} performance in different physiological settings, nine profiles were defined, as shown in Figure \ref{fig:vasculature}. 
The intuition behind defining the profiles lies in the fact that they feature different combinations of weight, height, and activity levels, representing the diverse situations the model may encounter in real-life scenarios. 
On the one hand, some of these profiles test the effect of modifying features individually, i.e., ``tall'' and ``short'' for the height, ``heavy'' and ``light'' for the weight, and ``active'' and ``inactive'' for the blood speed scale. On the other, the ``big'' and ``small'' profiles allow for assessing the model's performance when several attributes are modified simultaneously.

As the y-axis is assumed to represent the ``vertical axis'' of the nanodevice in the bloodstream, a modification in their height can be easily approximated by simply multiplying the Y coordinate by a height scaling ratio. 
When modifying the subject's weight, accurately estimating its impact on flow-guided localization is significantly more challenging, given that an actual increase in a person's weight does not always cause the same body regions to increase in size. 
Regardless, scaling the subject's weight in these experiments aims to determine whether a non-vertical change in the subject's vessels' positions affects the localization's performance. 
Therefore, the weight scaling of the simulation's subject for our experiments is considerably simplified, and it merely consists of a constant scaling of the X and Z coordinates, similar to the height modification. 
For this, we approximate the body's shape to a cylinder assumed to have uniform density, such that an increase of the weight is directly proportional to an increase in the cylinder's volume. 
Since the weight and height scaling ratios are known, the radius scaling proportion $k$ can be obtained based on the cylinder volume equation, resulting in the square root of the product of the weight ratio and the inverse height ratio.

Changing the coordinates alone would not impact the performance of the studied model, since the coordinates are not used as features themselves. Instead, the circulation time is used to infer the location of the detected events. In order to estimate how the circulation time varies based on the modified blood vessels, the time differences between circulation moments are scaled accordingly, and these are then added cumulatively for each simulation. 
To scale these, two cases are assumed for each pair of reported circulation times: i) both times are reported in the same region, and ii) each time is reported in a different region.

The first case is significantly simpler, as it merely requires the time difference to be scaled based on the length ratio of the current vessel with the new coordinates and the original coordinates (assuming a linear variation).
The second case, however, is more complex, since the vessel length ratio of each region will be different, and the exact original speed of the nanodevices in each region is unknown (only the distance between both coordinates and the total time difference are known). 
For this reason, the speed is estimated to be constant during the region transition, and the speed is calculated as the original traversed distance divided by the original time difference, which is then used to compute the new time difference using the scaled traversed distance. 
While this is not completely accurate (as the flow speed is different depending on the region type), most of the reported circulation times fall within the first case, and this is simply an approximate estimation of a real case, so we consider this as a negligible inaccuracy.

As a result, the final circulation time difference is the scaled time  difference divided by the blood speed scale (representing the activity level).
Using these modifications, the circulation times of each simulation are processed by the pipeline, which uses them to extract the anchor features explained in Section~\ref{sec:gnn_architecture}. 
These, along with the weight, the height and the activity level, are used to build the input graph, which is fed to the \ac{GNN} and outputs a vector of probabilities of an event being located in each region, with the highest probability representing the estimated one.

%!TEX root = main.tex

\section{Evaluation Methodology and Results}
\label{sec:results} 

\subsection{Evaluation Methodology}

In our evaluation, we utilize a framework for objective performance evaluation of \ac{THz}-based flow-guided localization from~\cite{lopez2023toward}. 
The framework combines two simulators, i.e., \ac{BVS}~\cite{bvs} and TeraSim~\cite{HOSSAIN2019100004}. 
\ac{BVS} offers a bloodstream simulation environment for flow-guided nanodevices. 
Terasim is an ns3-based platform for modeling THz communication networks, both an macro and nanoscale.
The combined framework offers the possibility of simulating the movement of flow-guided nanodevices across the bloodstream, where they are envisioned to detect events of diagnostic interest and communicate the sensing indicators to the on-body anchors.
This framework uses a set of input parameters defining an evaluation scenario, as shown in Table~\ref{tab:sim_params}. 
The inputs are envisioned to be passed to the framework for generating raw data for streamlined evaluation of a given flow-guided localization approach for the assumed scenario, resulting in a performance benchmark.

Along with these parameters, the framework is given information about the number and locations of anchors (nb., one anchor in the proximity of the heart is assumed in this work), which are assumed to be static entities and feature sufficient energy for continuous operation.
The nanodevices are assumed to be mobile energy-harvesting entities within the bloodstream. 
Then, BVS is invoked for generating the mobility profiles of the nanodevices in the bloodstream within a simulation time-frame. 
At each BVS-originating location of a nanodevice, the nanodevice is assumed to carry out a sensing/actuation task. 
This is followed by attempting at communicating the data (i.e., a binary event detection indicator, iteration time since last communication with the anchor, and the nanodevice identifier) to the anchor at \ac{THz} frequencies, which is then used by the \ac{GNN}. 
Table~\ref{tab:sim_params} summarizes the main simulation parameters, with the rest using the same values as the original \ac{GNN} model~\cite{bartra2023graph} for providing meaningful comparison with the baseline.

Our evaluation is carried out by hard-coding the location of an event of diagnostic interest in each of the 25 cardiovascular regions of the bloodstream representing organs, limbs, and head.
We obtained the simulated raw data for two event locations per region.
This procedure was repeated for the nine defined profiles, resulting in 450 simulated raw data profiles to be fed into the both of the considered \ac{GNN} solutions for benchmarking purposes.

For evaluating the performance of the considered flow-guided localization,~\cite{lopez2023toward} proposed a set of standardized performance metrics. 
These metrics are the region accuracy, the point accuracy, the reliability, and the nanodevice energy consumption. 
Since~\cite{bartra2023graph} has demonstrated the rather high reliability of the proposed \ac{GNN} model, as well as due to the fact that the nanodevices' energy consumption is related to their energy harvesting capabilities, which are unchanged across approaches, we focus on the region and point accuracies as the performance metrics relevant for this study. 

Region accuracy represents the fraction of correctly estimated regions over the total number of estimations, while the point accuracy corresponds to the amplitude of localization errors obtained as the Euclidean distance between the predicted and ground truth location coordinates.
The point accuracy distributions are captured across the considered 25 region types, which are depicted using regular box-plots. 
This metric allows for the performance analysis based on the prediction's spatial location instead of just the strict region, so predicting a neighboring region to the ground truth region yields better point accuracy than predicting a distant region. 

\begin{table*}[!htb]
\begin{minipage}{.29\linewidth}
\centering
\caption{Simulation parameters}
\label{tab:sim_params}
    % \medskip
    \vspace{-1mm}
\begin{tabular}{|l|c|} 
\toprule
\makecell{ \textbf{Parameter}}  &  \textbf{Value}  \\   
\midrule
Number of anchors             & 1       \\ \hline
Simulation time               & 1100~s  \\ \hline
Number of nanodevices         & 64      \\ \hline
Event sampling granularity    & 3~Hz    \\ \hline
Simulations per region        & 2       \\ \hline
Event detection threshold     & 1~cm    \\ \hline
Frequency                     & 1~THz \\ 
\bottomrule
\end{tabular}
\end{minipage}\hfill
\begin{minipage}{.7\linewidth}
    \centering
\caption{GNN hyperparameter search space and the best hyperparameters for each design}
\vspace{-1mm}
\label{tab:gnn_search_space}
    % \medskip
\begin{tabular}{|l|l|l|l|l|l|} 
    \toprule
    \makecell{ \textbf{Parameter}}  & \textbf{Search space}   & \textbf{Graph a}  & \textbf{Graph b}  & \textbf{Graph c}  & \textbf{Baseline}\\
    \midrule
Hidden channels (HC)    & \{16, 32, ... , 512\}  & 64                & 64                & 512               & 512\\ \hline
HGT heads (HHGT)          & \{1, 2, 4, 8\}                 & 2                 & 4                 & 8                 & 1  \\ \hline
GAT heads (HGAT)          & \{1, 2, 3\}                    & 1                 & 1                 & 1                 & 1  \\ \hline
HGT layers (HGTL)         & \{0, 1, 2, 3\}                 & 3                 & 2                 & 3                 & 2  \\ \hline
First layers (FL)       & \{0, 1, 2, 3\}                 & 1                 & 4                 & 3                 & 2  \\ \hline
Last layers (LL)        & \{0, 1, 2, 3\}                 & 3                 & 4                 & 4                 & 3  \\ \hline
Convolution type (CT)   & \{GAT, GCN\}                   & GAT               & GCN               & GCN               & GAT\\ \hline
Learning rate (LR)      & {[}1E-5, 0.01{]}               & 5e-3              & 5e-4              & 2e-3              & 4e-4\\ \hline
Weight decay (WD)       & {[}1E-6, 5E-3{]}               & 9e-6              & 1e-3              & 8e-5              & 8e-5\\ \hline
Max. gradient norm (MN) & {[}0.5, 5{]}                   & 4.5               & 4.4               & 3.4               & 3.9 \\ 
    \bottomrule
\end{tabular}
\end{minipage}
\end{table*}

\subsection{Hyperparameter Tuning}
We trained the considered models using grid search with the hyperparameter configuration search space shown in Table~\ref{tab:gnn_search_space}.
These grid search runs were executed separately for each of the three extended graph designs for finding their optimal configurations. 
The graph design with the best validation accuracy after the hyperparameter tuning was then used in the evaluation on each of the nine profiles under study and compared to the baseline.

\begin{figure*}[!t]
\centering
\vspace{-2mm}
\subfigure[Baseline GNN]{
\includegraphics[width=0.47\linewidth]{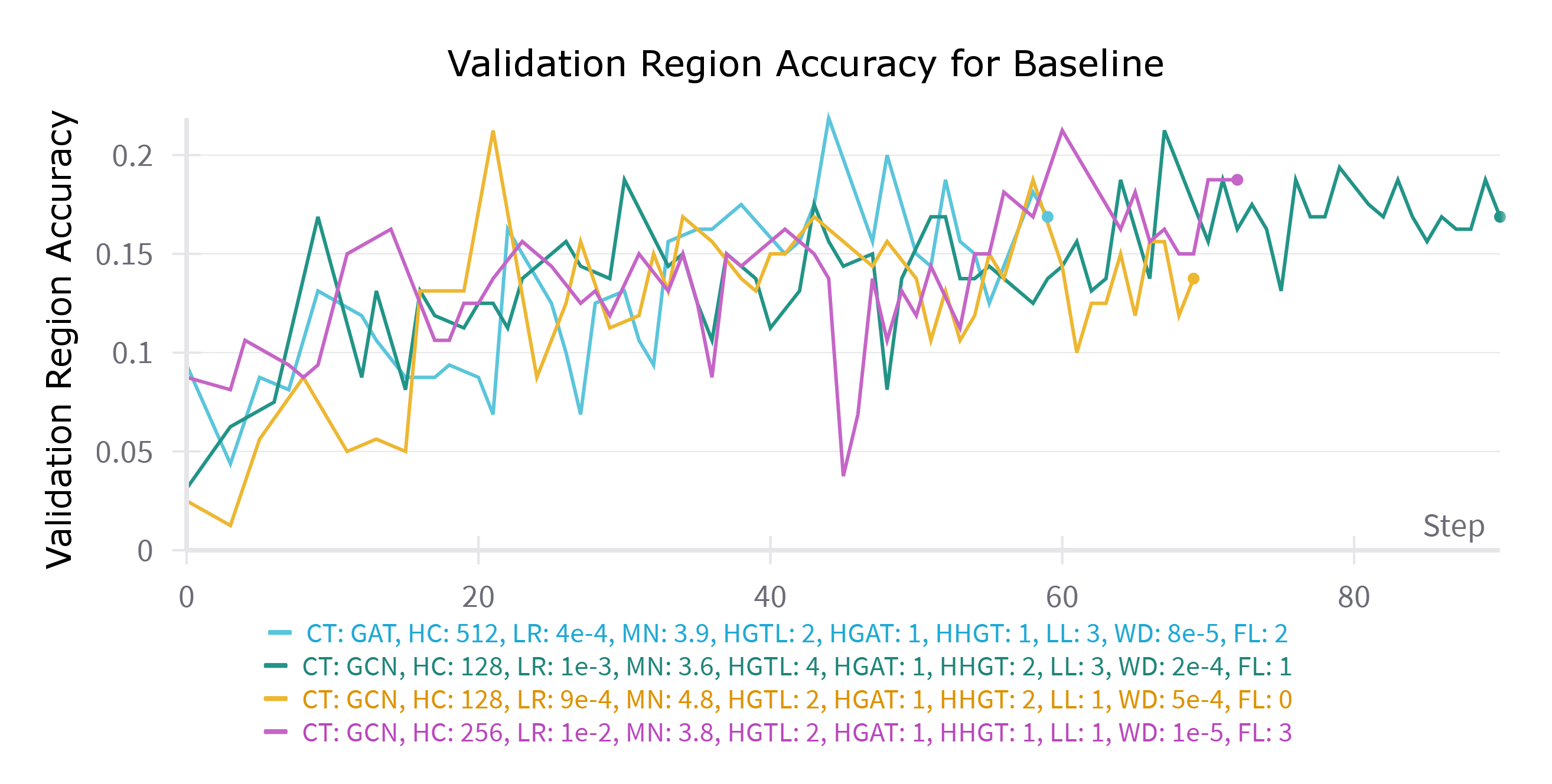}}
\subfigure[Extended GNN c, with probability edges and inverted master features]{
\includegraphics[width=0.47\linewidth]{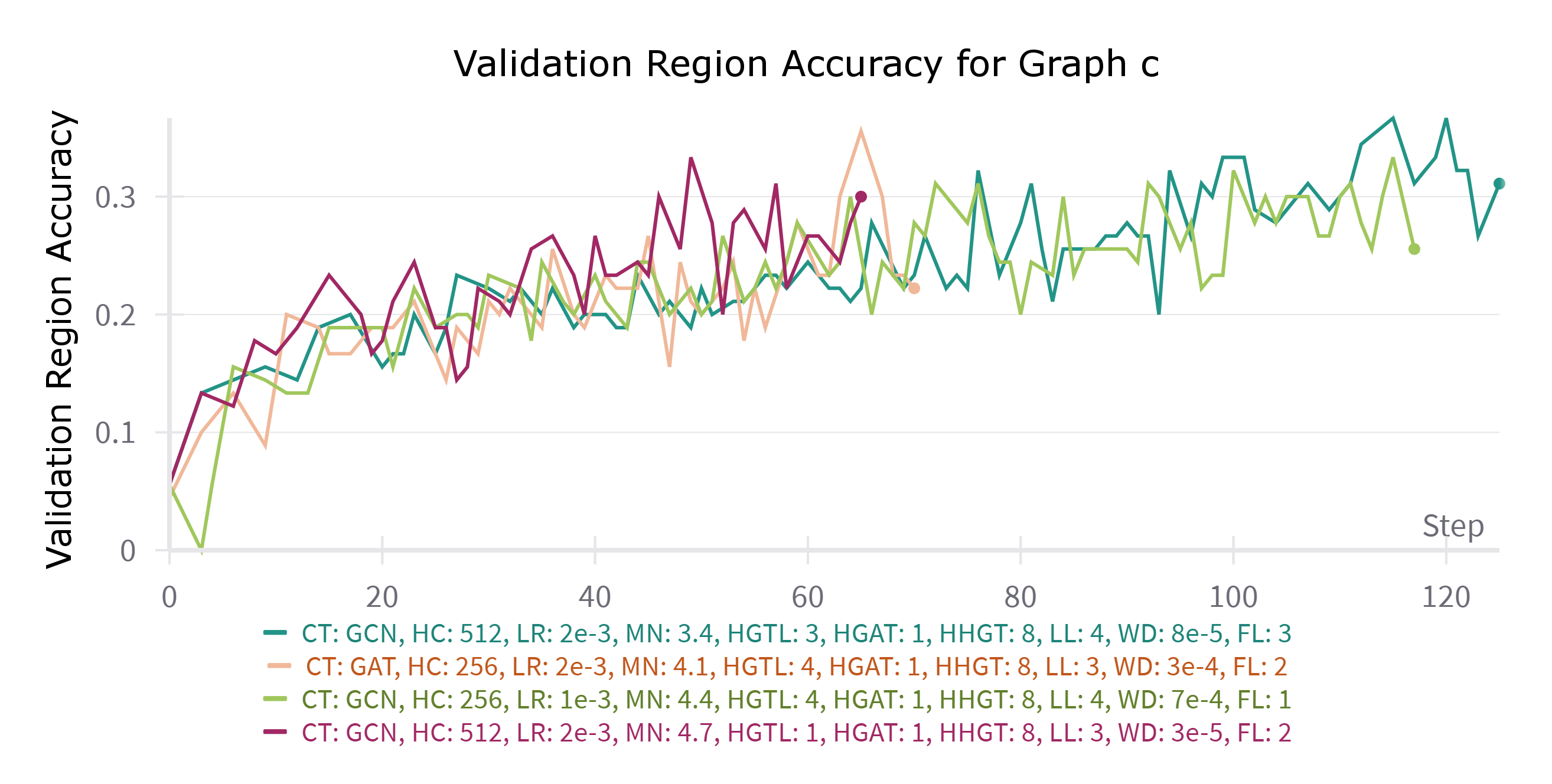}}
\vspace{-3mm}
\caption{A snapshot of the hyperparameter tuning process}
\label{fig:hyperparameter_search}
\vspace{-2mm}
\end{figure*} 

Figure~\ref{fig:hyperparameter_search} depicts a snapshot of the hyperparameter tuning process for the best performing extended \ac{GNN} design (i.e., GNN design c, featuring both the probability edges and inverted master features), as well as for the baseline.
The results are depicted for the best four runs only for visibility purposes. 
As visible, the three designs yielded a best validation accuracy of 30\%, 32.22\% and 36.67\%, respectively, showing that the best option is the extended design c, with the inclusion of both probability edges and inverted features. 
This is to be expected, as both of these changes were expected to provide useful additional information to the model.
In addition, the baseline model was trained on the original profile and using original graph design without a master node, probability edges, or inverted features, achieving the validation accuracy of 21.88\%.

\subsection{Evaluation Results}

\begin{figure*}
\centering
\begin{minipage}{.55\textwidth}
    \centering
    \includegraphics[width=\linewidth]{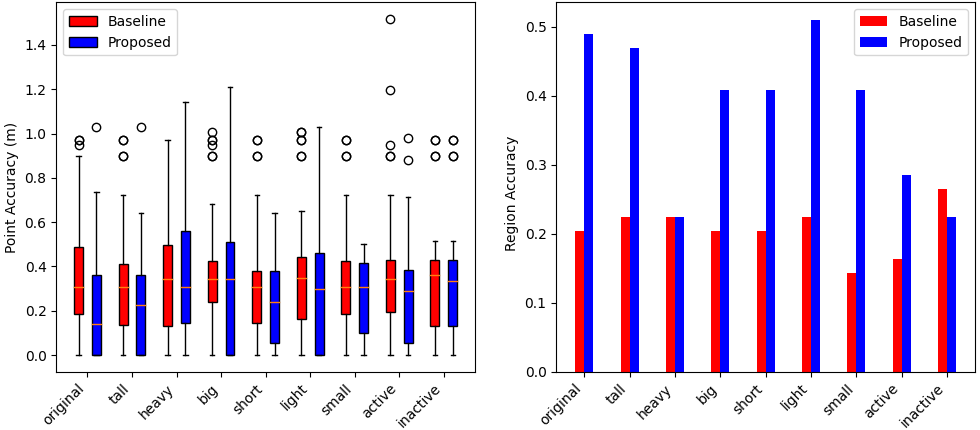}
    % \vspace{-3mm}
    \caption{Point (left) and region (right) accuracy scores}
    \vspace{-2mm}
    \label{fig:results_0}
\end{minipage}%
\begin{minipage}{.44\textwidth}
\subfigure[Original]{
\includegraphics[width=0.312\linewidth]{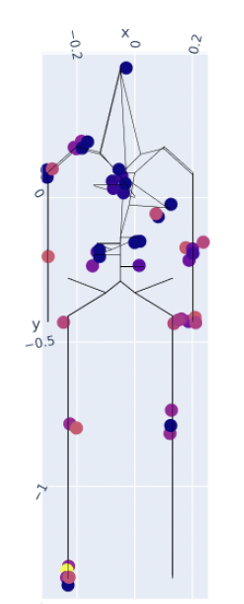}}
\subfigure[Light]{
\includegraphics[width=0.244\linewidth]{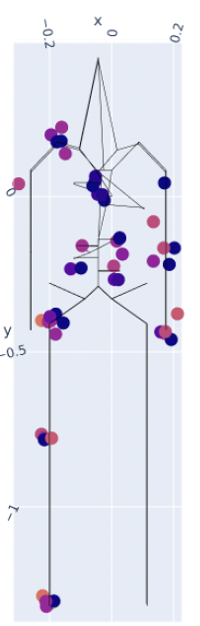}}
\subfigure[Heavy]{
\includegraphics[width=0.305\linewidth]{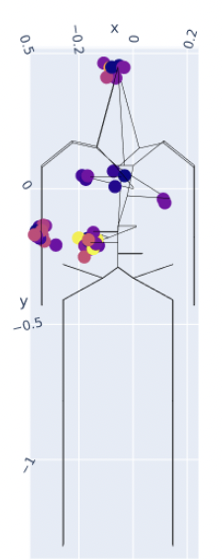}}
\subfigure{
\includegraphics[width=0.07\linewidth]{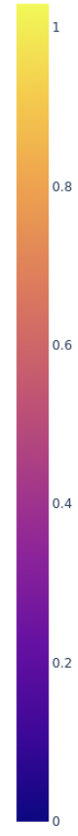}}
% \vspace{-4mm}
\caption{Point accuracy distribution for selected profiles}
\label{fig:point_distribution}
\end{minipage}
\vspace{-2mm}
\end{figure*}

\begin{figure*}[!t]
\centering
% \vspace{-4mm}
\subfigure[Original]{
\includegraphics[width=0.3\linewidth]{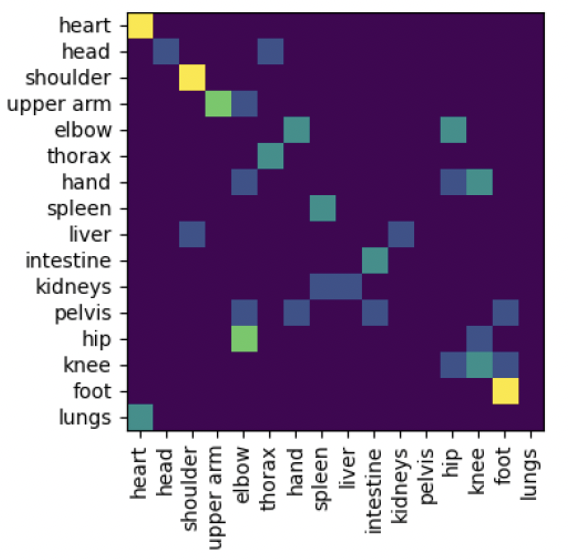}}
\subfigure[Light]{
\includegraphics[width=0.3\linewidth]{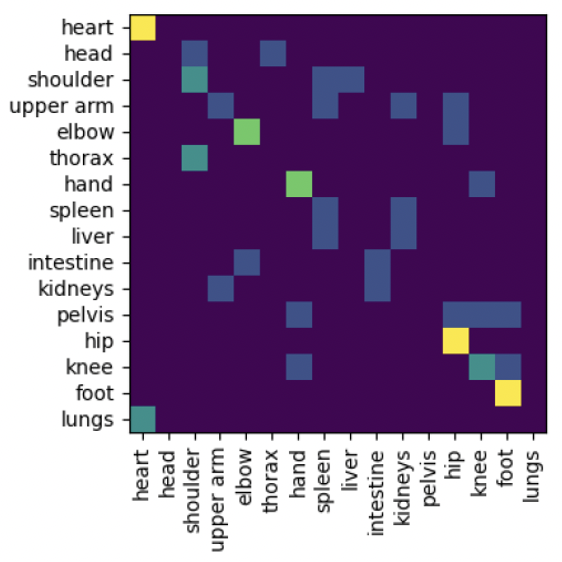}}
\subfigure[Heavy]{
\includegraphics[width=0.3\linewidth]{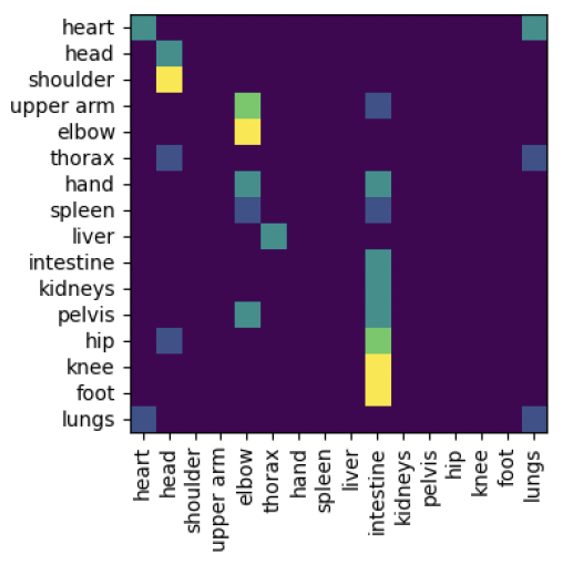}}
\vspace{-2mm}
\caption{Confusion matrices for selected profiles}
\label{fig:confusion_matrices}
\vspace{-3mm}
\end{figure*}

Figure~\ref{fig:results_0} depicts the point and region accuracies for each profile achieved by the considered baseline and best performing extended \ac{GNN} model. 
The region accuracy of the baseline is shown to decrease for the small and active profiles compared to the original, while for the other profiles it remains around 20\%. 
In contrast, the proposed model displays a significantly higher region accuracy for all except the heavy profile with equal accuracies and the inactive profile, where the proposed model performs slightly worse than the baseline. 
This is due to the time scaling dramatically modifying the circulation times, significantly more than in the other profiles, which makes them hardly distinguishable by both models.
Similarly, the low region accuracy in the heavy profile is contributed to the significant cylindric distortion along causing an unequal scaling of iteration times across regions.
This is demonstrated in Figures~\ref{fig:point_distribution} and~\ref{fig:confusion_matrices}, respectively depicting the spatial distributions of localization errors and confusion matrices of the selected profiles, both featuring significant performance distortion in the heavy profile.

% \begin{figure*}[!ht]
%     \centering
%     \includegraphics[width=0.65\linewidth]{Plots/performance_region2.png}
%     \vspace{-3mm}
%     \caption{Point (left) and region (right) accuracy scores}
%     \vspace{-3mm}
%     \label{fig:results_0}
% \end{figure*}

Moreover, the point accuracy distributions across models demonstrate that in cases where the proposed model does not surpass the baseline in terms of region accuracy, its point accuracy is either the same or higher than the one achieved by the baseline.
This suggests that the incorrectly predicted regions by our model are not far from the ground truth spatially.
Similarly, the small difference in the distribution of localization errors is observed across models in cases the proposed model substantially outperforms the baseline.
This can be attributed to the fact that the baseline, when distorted, predicts regions neighboring to the correct one.

%!TEX root = main.tex

\vspace{1mm}
\section{Conclusion}
Our investigation into individualized factors affecting flow-guided nanoscale localization reveals that individual differences in the bloodstream sizes or patients' activities can substantially impact localization accuracy.
Toward addressing this issue, we provide a pipeline for adapting the raw data for flow-guided localization based on physiological attributes. 
The enhancements introduced in this work are beneficial in terms of the \acf{GNN} model's adaptability, given that they are almost generally yield better performance than the reference model without adaptations.
In more general terms, these findings underscore the importance of tailoring localization strategies to individual bloodstreams, considering variations in height, weight, and heart rate. 
As flow-guided nanoscale localization advances, understanding and accommodating individual differences will be pivotal for realizing its full potential in medical applications.

While this study focused on the performance of our framework on the nine discrete patients' profiles, the proposed graph adaptations of the original \ac{GNN} are expected to support continuous adaptation of the physiological parameters by taking data from time windows when these values remain approximately constant. 
The performance of the \ac{GNN} model and proposed graph adaptations when continuously changing the physiological parameters such as weight and activity levels will be considered as a part of our future efforts. 
In addition, future work will focus on creating methods for further mitigation of temporally variable philological parameters by either improving the preprocessing of the input data or by enhancing the \ac{GNN} with new features, parameters, or architecture elements. 
Here we will primarily focus on the activity levels as the most dynamic of the considered physiological parameters.
Since we only use the raw data received by a single anchor in the proximity of the heart, our model is by-design unable to distinguish between left and right counterparts on the body (e.g., limbs). 
Using multiple anchors is expected to further refine the proposed model accuracy, but potentially also its adaptability, which might render it useful for currently challenging body profiles.

\vspace{-1mm}
 \bibliographystyle{IEEEtran} 
 \bibliography{bibliography}

\end{document}